\documentclass{bmvc2k}
\usepackage{graphicx}
\usepackage{algorithm}
\usepackage{algpseudocode}
\newcommand{\wenxuan}[1]{\textcolor{black}{{#1}}}
\title{Face Alignment Assisted by Head Pose Estimation}

\addauthor{Heng Yang}{heng.yang@cl.cam.ac.uk}{1}
\addauthor{Wenxuan Mou}{w.mou@qmul.ac.uk}{2}
\addauthor{Yichi Zhang}{yichizhang@fas.harvard.edu}{3}
\addauthor{Ioannis Patras}{i.patras@qmul.ac.uk}{2}
\addauthor{Hatice Gunes}{h.gunes@qmul.ac.uk }{2}
\addauthor{Peter Robinson}{peter.robinson@cl.cam.ac.uk}{1}

\addinstitution{
 Computer Laboratory\\
 University of Cambridge\\
 Cambridge, UK
}
\addinstitution{
School of EECS\\
 Queen Mary University of London\\
 London, UK
}
\addinstitution{
Faculty of Arts \& Sciences\\
 Harvard University\\
 Cambridge, MA, US
}

\runninghead{Yang, Mou, Zhang et al.}{Face alignment assisted by Head Pose Estimation}

\begin{document}

\maketitle

\begin{abstract}
In this paper we propose supervised initialisation scheme for cascaded face alignment based on explicit head pose estimation.  We first investigate the failure cases of most state of the art face alignment approaches and observe that these failures often share one common global property, i.e. the head pose variation is usually large. Inspired by this, we propose a deep convolutional network model for reliable and accurate head pose estimation. \wenxuan{Instead} of using a mean face shape, or randomly selected shapes for cascaded face alignment initialisation, we propose two schemes for \wenxuan{generating initialisation}: the first one relies on projecting a mean 3D face shape (represented by 3D facial landmarks) onto 2D image under the estimated head pose; the second one searches nearest neighbour shapes from \wenxuan{a} training set according to head pose distance. By doing so, the initialisation \wenxuan{gets} closer to the actual shape, which enhances \wenxuan{the possibility of convergence} and in turn improves the face alignment performance. We demonstrate the proposed method on the benchmark 300W dataset and show very competitive performance in both head pose estimation and face alignment. 
\end{abstract}

\section{Introduction}
Both head pose estimation and face alignment have been well studied in recent years given their wide application in human computer interaction, avatar animation, \wenxuan{and} face recognition/verification. These two problems are very correlated and  putting them together will enable mutual benefits. Head pose estimation from 2D images remains a challenging problem due to the high diversity of face images \cite{haj2012partial, murphy2009head}. Recent methods \cite{fanelli2011real} attempt to estimate the head pose by using depth data. On the contrary, face alignment has made significant progress and several methods \cite{cfaneccv2014,asthanaincremental,renface,xiong2013supervised} have reported good performance on images \textit{in the wild}. However, they also show some failures. When we look into their failures cases, we find that those samples share one significant property, i.e., the head (face) in such images is usually rotated from frontal pose in big angles. 

\begin{figure}
\includegraphics[trim =0.0cm 0.0cm 0.0cm 0.0cm, clip = true, width=0.95\textwidth,height=0.33\textwidth]{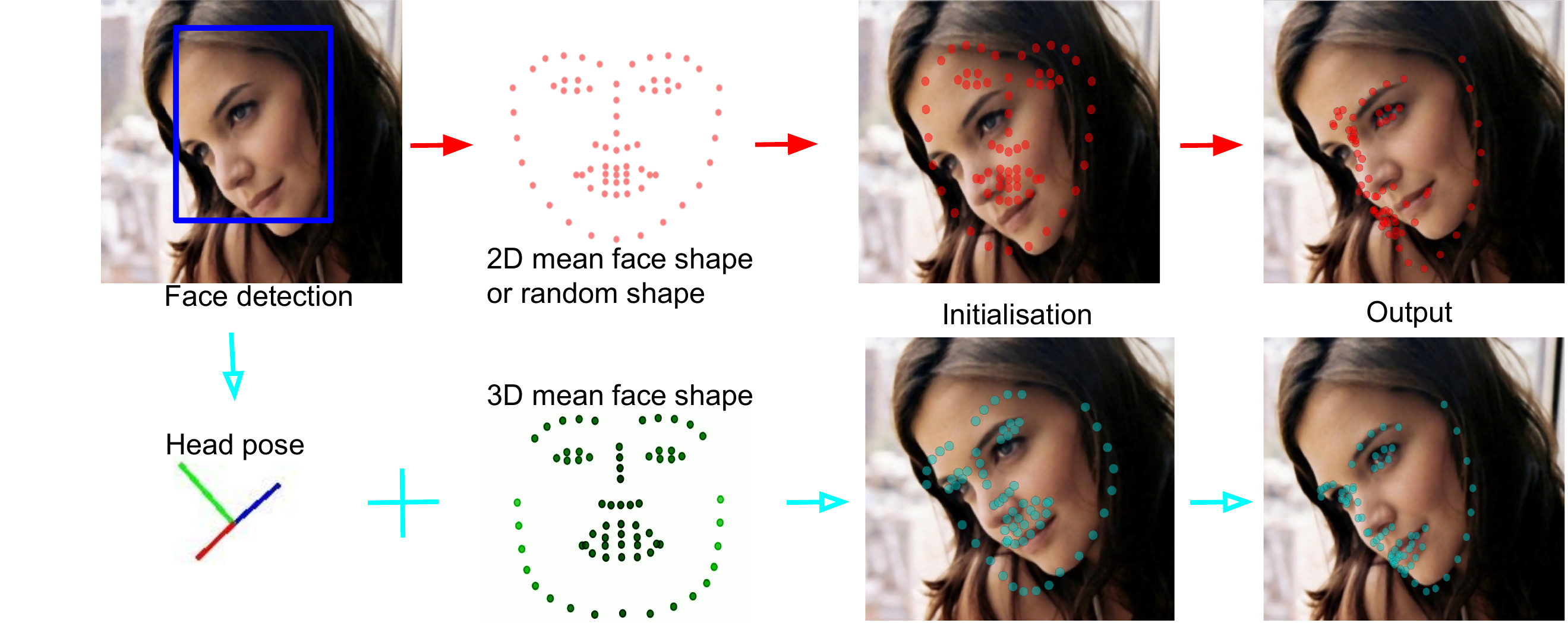}
\label{fig:illustration}
\caption{Our proposed head pose based cascaded face alignment procedure (path in \textcolor{cyan}{cyan} color) vs. conventional cascaded face alignment procedure (path in \textcolor{red}{red} color). }
\end{figure}

\wenxuan{The best performing face alignment methods proposed in recent years (\cite{xiong2013supervised}, \cite{asthanaincremental} and \cite{cfaneccv2014})} also share a similar cascaded pose regression framework, i.e., face alignment starts from a raw shape (a vector representation of the landmark locations), and updates the shape in a coarse to fine manner. The methods in this framework are usually initialisation dependent. Therefore, the final output of one cascaded face alignment system might change if a different initialisation is \wenxuan{provided} to the same \wenxuan{input} image. Moreover, each model has a convergence radius, i.e., if the initialisation lies within the range of the actual shape, the model will be able to output a reasonable alignment result, otherwise it might lead the shape to a wrong location, as shown in Fig.~\ref{fig:illustration}.  The methods like \cite{xiong2013supervised,asthanaincremental} \wenxuan{perform initialisation} using a mean shape within the face bounding box or from a randomly selected shape from training set. \wenxuan{There} is no guarantee the initialisation lies within the convergence radius, especially when head pose variation is large. 

In this paper, we aim to address \wenxuan{the} above discussed problems and make cascaded face alignment perform better under large head pose variations. The difference between our proposed method and the conventional cascaded method procedure is illustrated in Fig.~\ref{fig:illustration}.  In contrast to using mean shape or random shapes for initialisation by other methods, our proposed method aims to produce better initialisation schemes for cascaded face alignment based on explicit head pose estimation.  This is motivated by two facts: 1) most current methods fail on face images with large head pose variation-as we will demonstrate later; 2) most recent face alignment methods work in a cascaded fashion and perform initialisation with mean shape. More specifically,
we first estimate the head pose using a deep Convolutional Network (ConvNet) directly from face image. Given the estimated head pose, we propose two schemes of producing the initialisations. The first scheme projects a canonical 3D face shape under the estimated head pose to the detected face bounding box. The second scheme searches shape(s) for initialisation from the training set by nearest neighbour method in the head pose space. We build on our proposed scheme on the Robust Cascaded Pose Regression (RCPR) to demonstrate the effectiveness of supervised initialisation. We note that the proposed initialisation scheme can be naturally applied to any other cascaded face alignment. In summary, we make the following contributions:
\begin{itemize}
\item We investigate the failure cases of several state of the art face alignment approaches and find \wenxuan{that} the head pose variation is a common issue across those methods. 
\item Based on the above observation, we propose a ConvNet framework for explicit head pose estimation. It is able to achieve an accuracy of 4$^{\circ}$ absolute mean error of head pose estimation for face images acquired in unconstrained environment. 
\item We  propose two initialisation schemes based on reliable head pose estimation. They \wenxuan{enable} face alignment method (RCPR) perform better and reduce  large head pose failures by 50\% when using only one initialisation. 
\end{itemize}
To summarise, we propose  better initialisation schemes based on explicit head pose estimation for cascaded face alignment, to improve the performance, especially in the case of large head pose variation.

\section{Related Work}
Face alignment has made considerable progress in the past years and a large number of methods have been proposed. \wenxuan{There} are two different sources of information typically used for face alignment: face appearance (i.e., texture of the face image) and the shape information. Based on how the spatial shape information is used, the methods are usually categorized into local-based methods and holistic-based methods. The methods in the former category usually rely on discriminative local detection and use explicit deformable shape models to regularize the local outputs while the methods in the latter category directly regress the shape (the representation of the facial landmarks) in a holistic way, i.e. the shape and appearance are modelled together. 

\subsection{Local-based methods}
Local based methods usually consist of two parts. One is for local facial feature detection, which is also called local experts and the other is for spatial shape models. The former describes how image around each facial landmark looks like in terms of local intensity or color patterns while the latter describes how face shape, that is the relative location of the face parts, varies. This captures variations such as wide forehead, narrow eyes, long nose etc. 

There are three types of local feature detection. (1) Classification methods include Support Vector Machine (SVM) classifier \cite{rapp2011multiple,belhumeur2011localizing} based on various image features such as Gabor \cite{vukadinovic2005fully}, SIFT \cite{lowe2004distinctive,xiong2013supervised}, HOG \cite{yanlearn} and multichannel correlation filter responses \cite{Kiani_2013_ICCV}. (2) Regression-based approaches are also widely used. For instance, Support Vector Regressors (SVRs) are used in \cite{martinez2012local} with a probabilistic MRF-based shape model and Continuous Conditional Neural Fields (CCNF) are used in \cite{baltruvsaitis2014continuous}. (3) Voting-based approaches are also introduced in recent years, including regression forests based voting methods \cite{cootesECCV2012,dantone2012real,yangiccv2013} and exemplar based voting methods \cite{smithnonparametric,shen2013detecting}. 

One typical shape model is the Constrained Local Model (CLM) \cite{cristinacce2006feature}. The CLM steps can be summarised as follows: first, sample a region from the image around the current estimate and project it into a reference frame; second, for each point, generate a  ``response image" giving a cost for having the point at each pixel; third, searching for a combination of points which optimises the total cost, by manipulating the statistical shape model parameters. The methods built on CLM mainly differ from each other in terms of local experts, for instance CCNF in \cite{baltruvsaitis2014continuous} and the Discriminative Response Map Fitting (DRMF) in \cite{asthana2013robust}. There are many other local based methods either using CLM or other models such as RANSAC in \cite{belhumeur2011localizing}, graph-matching in \cite{Zhou_2013_ICCV}, Gaussian Newton Deformable Part Model (GNDPM) \cite{tzimiropoulos2014gauss} and mixture of trees \cite{devacvpr2012face}. 

\subsection{Holistic-based methods}
\begin{table*}[!hbtp]
\footnotesize
\setlength{\tabcolsep}{1.5pt}
\centering
\caption{Holistic methods and their properties.}
\label{tab::holisticmethods}
\begin{tabular}{lcccccc}
\hline
Methods        & SDM \cite{xiong2013supervised}       & RCPR \cite{burgos2013robust}         &  IFA \cite{asthanaincremental} & LBF  \cite{renface}        & CFAN \cite{cfaneccv2014}        & TCDCN \cite{zhang2014facial} \\
initialisation & mean pose & random       & mean pose     & mean pose & supervised & supervised  \\
features       & SIFT      & pixel     &HOG    & pixel         & auto-encoder & ConvNet feature\\
regressor      & linear regression    & random ferns & linear regression & random forests & linear regression&   ConvNet    \\
\hline
\end{tabular}
\end{table*}
Holistic methods have gained high popularity in recent years and most of them work in a cascaded way like SDM \cite{xiong2013supervised}  and RCPR \cite{burgos2013robust}. We list very recent holistic methods as well as their properties in Table~\ref{tab::holisticmethods}. The methods following the  cascaded framework differ from each other mainly in three aspects. First, how to set up the initial shape; Second, how to calculate the shape-indexed features; Third, what type of regressor is applied at each iteration. \wenxuan{For} initialisation, there are mainly three strategies are proposed in literature: random, mean pose, and supervised.  
In order to make it less sensitive to initialisation, previous approaches such as \cite{suncvpr2012, burgos2013robust} propose to run multiple different initialisations and pick the median of all the predictions as the final output. Each initialisation is treated \wenxuan{independently} way until the output is calculated. However, such a strategy has several issues, first the theoretical support \wenxuan{for} selecting the median value is not well understood; second, there is no guidance on how to choose the multiple initialisations; third, using multiple initialisations is computationally expensive.  A similar supervised initialisation scheme was proposed in \cite{yang2015robust} where the initialisation shapes were selected by using an additional regression forest model for sparse facial landmarks estimation. A recent work \cite{yang2015mirror} proposed a re-initialisation scheme based on mirrorability to improve the face alignment performance. 
 
\section{Data preparation}
In this section we describe how the data is prepared in order to support our further discussion. More specifically, we discuss how we provide ground truth head pose and face bounding boxes from different face detectors for the benchmark dataset. 

We use face image data from the benchmark face alignment in the wild dataset, 300W \cite{sagonas300}. Since their testing samples are not publicly available, we follow the partition of recent methods \cite{renface} to set up the experiments. More specifically, we use face images from  AFW \cite{devacvpr2012face}, HELEN \cite{tan2009enhanced}, LFPW \cite{belhumeur2011localizing} and iBug \cite{sagonas300}, which include 3148 training images and 689 test images in total. 3148 training images are from AFW (337 images), HELEN training set (2000 images) and LFPW training set (811 images), and 689 test images are from HELEN test set (330 images), LFPW test set (224 images) and iBug (135 images).

It is intractable to get the ground truth 3D head pose for face images collected in unconstrained conditions. In order to generate reasonable head pose (Pitch, Yaw and Roll) values, we use the pose estimator provided \wenxuan{by} Supervised Descent Method (SDM) \cite{xiong2013supervised}. \wenxuan{Note} that, when calculating the head pose, we feed the ground truth facial landmark locations instead of using the detected landmarks. Technically, head pose is estimated by solving the projection function from an average 3D face model (49 3D points) to the \wenxuan{input} image\wenxuan{,} given the 3D to 2D correspondences. We also use the 3D head pose estimator provided by \cite{asthana2013robust} for head pose calculation for evaluating the results. It produces very similar results to \cite{xiong2013supervised}. We calculate the head pose for all images in 300W. 

The benchmark dataset only provides two types of face bounding boxes: one is the ground truth bounding box calculated as the tight box of the annotated facial landmarks; the other is the detection results from model of \cite{devacvpr2012face}, which is quite similar to the ground truth face bounding box. However, several models like SDM \cite{xiong2013supervised} and \wenxuan{RCPR} \cite{burgos2013robust} are trained with different face bounding boxes, thus their performance deteriorates significantly \wenxuan{when} using the provided face bounding boxes. We therefore provide different face bounding boxes to the test images by employing Viola-Jones detector \cite{viola2001rapid} and HeadHunter detector \cite{mathias2014face} for fair comparison. For the \wenxuan{input} images on which \wenxuan{the} face detector fails we manually set reasonable bounding boxes.

\begin{figure}
\includegraphics[trim =2.0cm 2.0cm 2.0cm 2.0cm, clip = true, width=0.45\textwidth,height=0.25\textwidth]{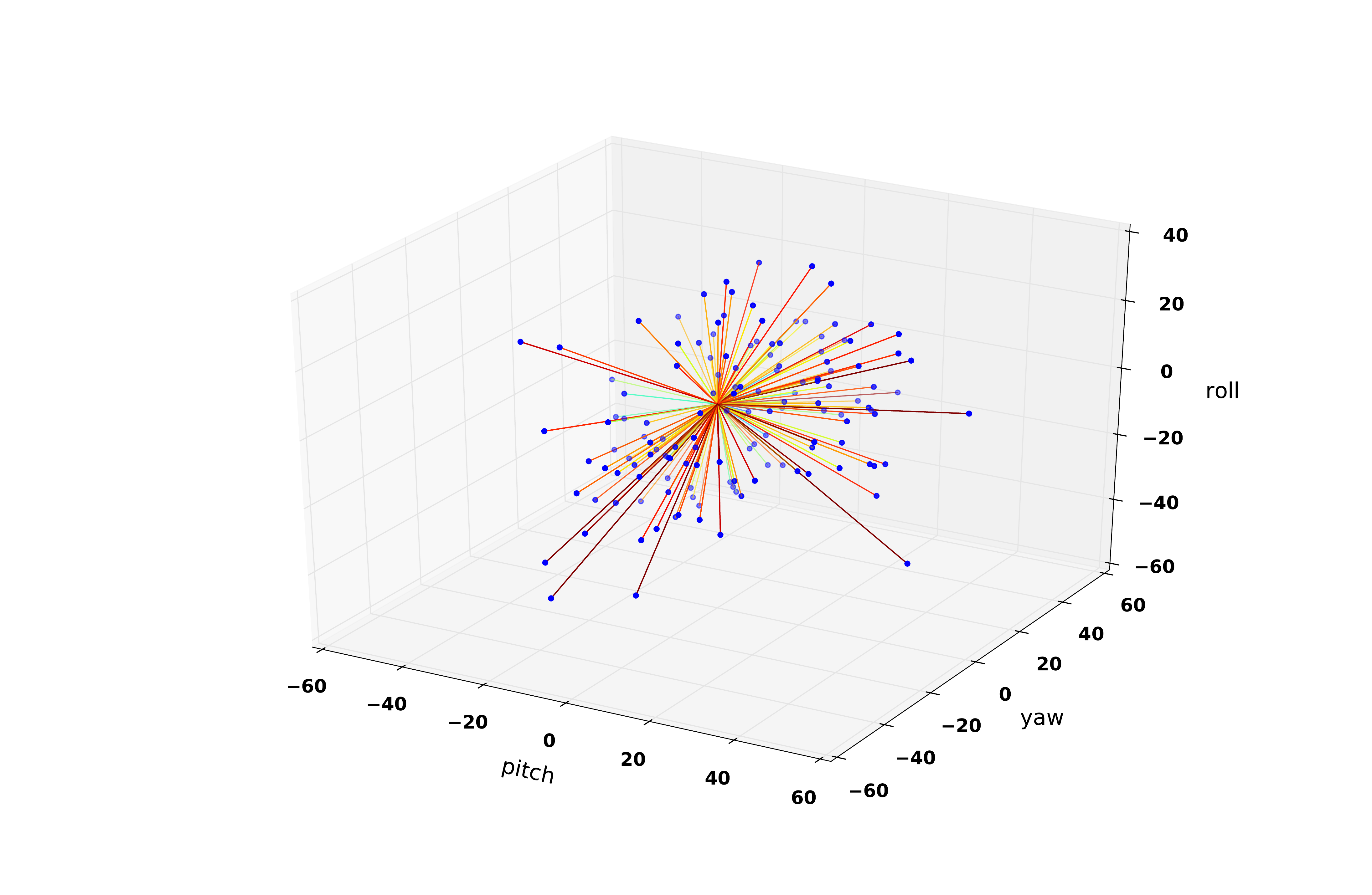}
\includegraphics[trim =0.0cm 0.0cm 0.0cm 0.0cm, clip = true, width=0.5\textwidth,height=0.25\textwidth]{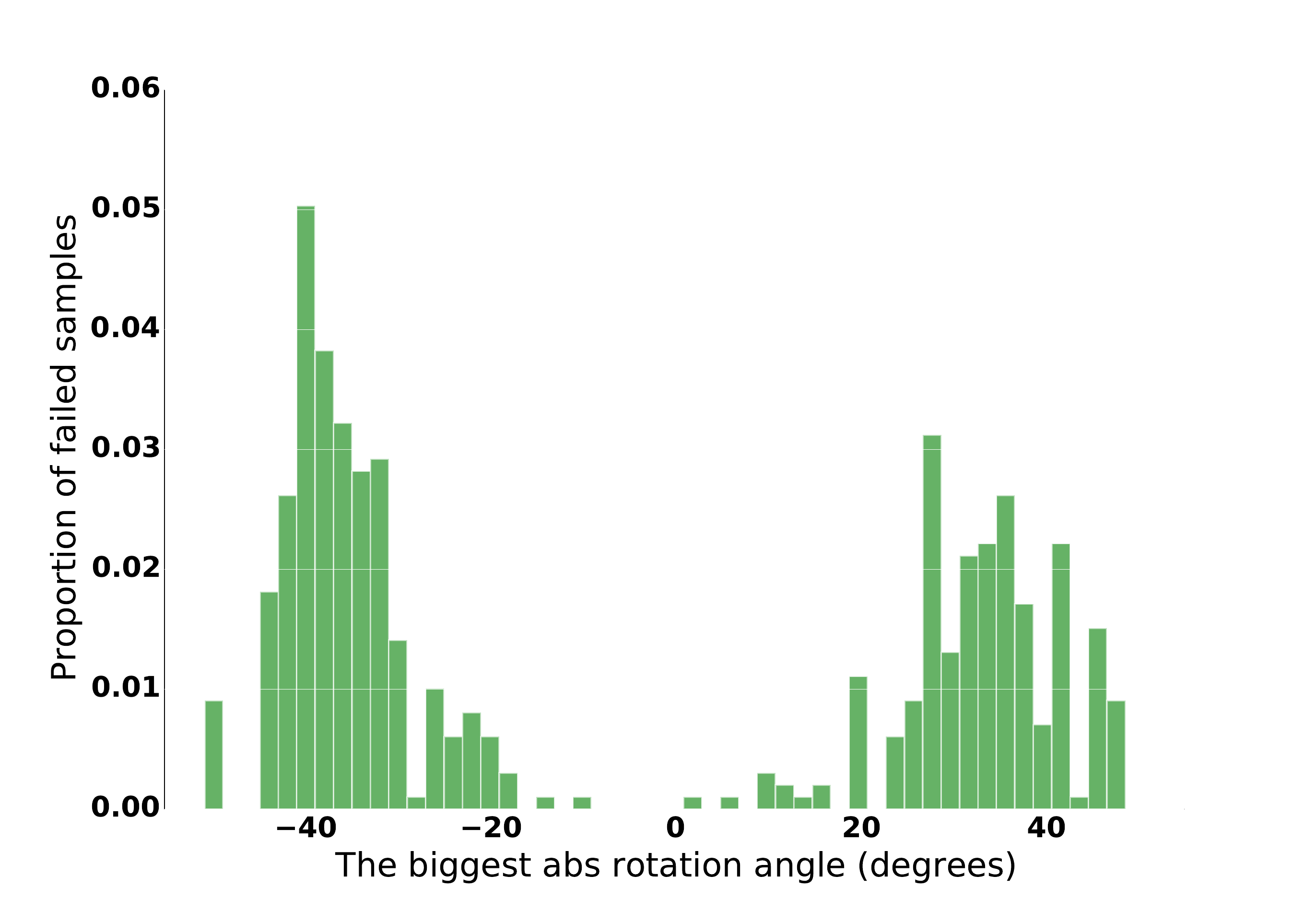}
\caption{Distribution of \wenxuan{the most} erroneous samples. }
\label{fig:toperror}
\end{figure}
\section{Method}
\subsection{Motivation}
\label{sec::motivation}
We first run several state of the art methods, including 6 holistic based methods (SDM \cite{xiong2013supervised}, IFA \cite{asthanaincremental}, LBF  \cite{renface}, CFAN \cite{cfaneccv2014}, TCDCN \cite{zhang2014facial}, RCPR \cite{burgos2013robust}) and 3 local based methods (GNDPM \cite{tzimiropoulos2014gauss}, DRMF \cite{asthana2013robust}, CCNF \cite{baltruvsaitis2014continuous}) given their good performance and availability of source \wenxuan{code}. For each method, we provide the \textit{best} type of  face bounding boxes in order to get the best performance. For each method, we select 50 difficult samples out of the 689 test samples that \wenxuan{provide} the biggest sample-wise alignment error. Then we plot their head poses in Fig.~\ref{fig:toperror} (left). As can be seen, most of the points are far away from the original point, i.e. they \wenxuan{have} big rotation angle(s). We further plot the histogram of the biggest absolute rotation angles of those samples in Fig.~\ref{fig:toperror} (right). The biggest absolute rotation angle is calculated as the one of the three directions with the biggest absolute value. As can be seen, those samples are distributed at big absolute angles. There are very few samples that \wenxuan{have} small rotation angles. Based on this observation, we can conclude that, large head pose rotation is one of the \wenxuan{main} factors that make most of the current face alignments fail. Based on this fact, we develop a head pose based initialisation scheme for improving the performance of face alignment under \wenxuan{large} head pose variations. 
%
\begin{figure}
\begin{tabular}{ccc}
\includegraphics[trim =0.0cm 0.0cm .0cm 0.0cm, clip = true, width=0.95\textwidth,height=0.25\textwidth]{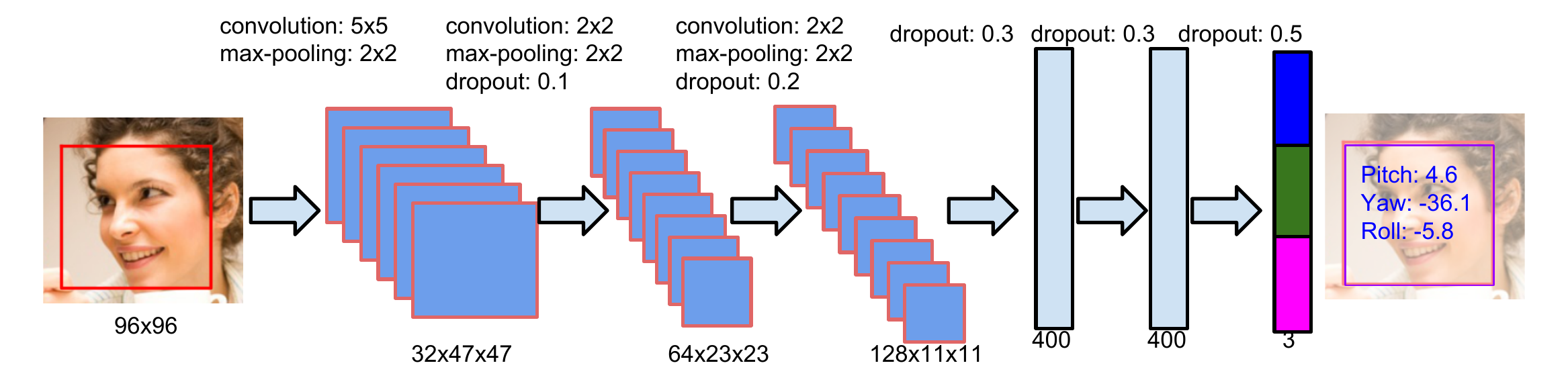}
\end{tabular}
\caption{ConvNet model for head pose estimation. }
\label{fig:headposenet}
\end{figure}
\subsection{Head Pose Estimation}
\label{sec::headposeestimation}
Giving the training data from 300W with augmented head pose annotation, we train a convolutional network (ConvNet) \cite{lecun1998gradient}  model for head pose estimation on the training set of 300W with 3148 images. The samples are augmented by 3 times with small permutations on the face bounding box. The ConvNet structure is shown is shown in Fig. \ref{fig:headposenet}. The input of the network is 96x96 gray-scale face image , normalised to the range between 0 and 1. The feature extraction stage contains three convolutional layers, three pooling layers, two fully connected layers and three drop-out layers. As we pose it as a regression problem, the output layer is 3x1 representing the head pose pitch, yaw and roll angle respectively. The angles are normalised between -1 and 1. We use Nesterov's Accelerated Gradient Descent (NAG) method \cite{sutskever2013importance} for parameter optimisation and we set the momentum to 0.9 and learning rate to 0.01. The training finishes in two hours on Tesla K40c GPU after around 1300 epochs, controlled by early-stop strategy. The learning curve is shown in Fig.~\ref{fig::headposeresult} (left).  The forward propagation of this network on GPU only takes 0.3ms per image on average.  
\subsection{Pose based Cascaded Face Alignment}
\subsubsection{General Cascaded Face Alignment}
In order to make this work stand alone, we first summarise the general framework of cascaded face alignment. Face shape is often represented as a vector of landmark locations, i.e., $S=(\mathrm{x}_1,...,\mathrm{x}_k,...,\mathrm{x}_K) \in\mathbf{R}^{2K}$, where $K$ is the number of landmarks. $\mathrm{x}_k \in \mathbf{R}^2$ is the 2D coordinates of the $k$-th landmark. Most of the current holistic-based method works in a coarse-to-fine fashion, i.e., shape estimation starts from an initial shape $S^0$ and progressively refines the shape by a cascade of $T$ regressors, $R^{1...T}$. Each regressor refines the shape by producing an update, $\Delta S$, which is added on the current shape estimate, that is,
\begin{equation}
S^t = S^{t-1} + \Delta S.
\end{equation}
The update $\Delta S$ returned from the regressor that takes the previous pose estimation and the image feature $I$ as inputs:
\begin{equation}
\Delta S = R^t(S^{t-1},I)
\end{equation}
An important aspect that differentiates this framework from the classic boosted approaches is the feature re-sampling process. More specifically, instead of using the fixed features, the input feature for regressor $R^t$ is calculated relative to the current pose estimation. This is often called pose-indexed feature as in \cite{dollar2010cascaded}. This introduces weak geometric invariance into the cascade process and shows good performance in practice. The CPR is summarized in Algorithm \ref{alg::algorithm1} \cite{dollar2010cascaded}.
\begin{algorithm}
\caption{Cascaded Pose Regression}
\label{alg::algorithm1}
\begin{algorithmic}[1]
    \Require{Image $I$, initial pose $S^0$} 
    \Ensure{Estimated pose $S^T$}
    \For {$t$=1 to $T$}
    \State $f^t = h^t(I,S^{t-1})$\Comment{Shape-indexed features} 
    \State $\Delta S = R^t(f^t)$\Comment{Apply regressor $R^t$}
    \State $S^t = S^{t-1}+\Delta S$\Comment{update pose}
     \EndFor
\end{algorithmic}
\end{algorithm}
\subsubsection{Head Pose based Cascaded Face Alignment}
In section \ref{sec::headposeestimation} we have presented how a ConvNet model can be used for head pose estimation. We propose two head pose based initialisation schemes for face alignment. One is based on an average 3D face shape projection and the other is based on nearest neighbour searching. 
\paragraph{Scheme 1: 3D face shape based initialisation} 
Given a 3D mean face shape, represented by 68 3D facial landmark locations,  as shown in Fig.~\ref{fig:illustration}, we first project this shape under the estimated head pose to a set of canonical 2D locations. More specifically we use constant translation and focus length in order to get a reasonable projection for all images. Then we re-scale the canonical 2D projection by the face bounding box scale of the test image to get the initialisation. We can represent the initialisation process by function $\mathcal{F}$ as follows. 
\begin{equation}
S_0 = \mathcal{F}(\theta,bb,\bar{S}^{3D})
\end{equation}
with $bb$ the face bounding box, $\bar{S}^{3D}$, the 3D mean face shape, $\theta$, the estimated head pose, which can be represented by:
\begin{equation}
\theta = \mathcal{G}(I, bb)
\end{equation}
where $\mathcal{G}$ is the deep convolutional model described in section \ref{sec::headposeestimation}. 

\paragraph{Scheme 2: Nearest Neighbour based initialisation} 
We propose a second scheme for head pose based initialisation by nearest neighbour search. Since we have provided the training samples with head pose information as well, we can easily search samples that are with similar head pose of a test sample. Then we calculate similarity transformation between two face bounding boxes in order to calculate the initialisation shape for the test sample. In this way, we can also provide $K$ initialisations by searching $k$-Nearest Neighbors from the training set.

Once we get a reliable initialisation (or several ones), we feed it to Algorithm \ref{alg::algorithm1} and apply the cascade of regressors in the same way to the baseline approach. In the case of the multiple initialisations, we calculate the output in a similar fashion to  \cite{burgos2013robust,suncvpr2012}, i.e., to pick up the median value of their estimations.  We build our proposed head pose based initialisation schemes on top of the popular Cascaded Pose Regression (CPR) method due to its simplicity and popularity. We train its recent variant Robust Cascaded Pose Regression (RCPR) \cite{burgos2013robust} model by using its new interpolated feature extraction, which is re-implemented by the author of \cite{yang2014face}. We do not use its full version as  occlusion status annotation is not available. We trained the baseline RCPR model on our 300W training set using Viola-Jones \cite{viola2001rapid} face detection. 20 random initialisations are used for data augmentation at the training time. 

\section{Evaluation}
 \begin{figure*}
\includegraphics[trim =0.0cm 0.0cm 0.0cm 0.0cm, clip = true, width=0.5\textwidth,height=0.25\textwidth]{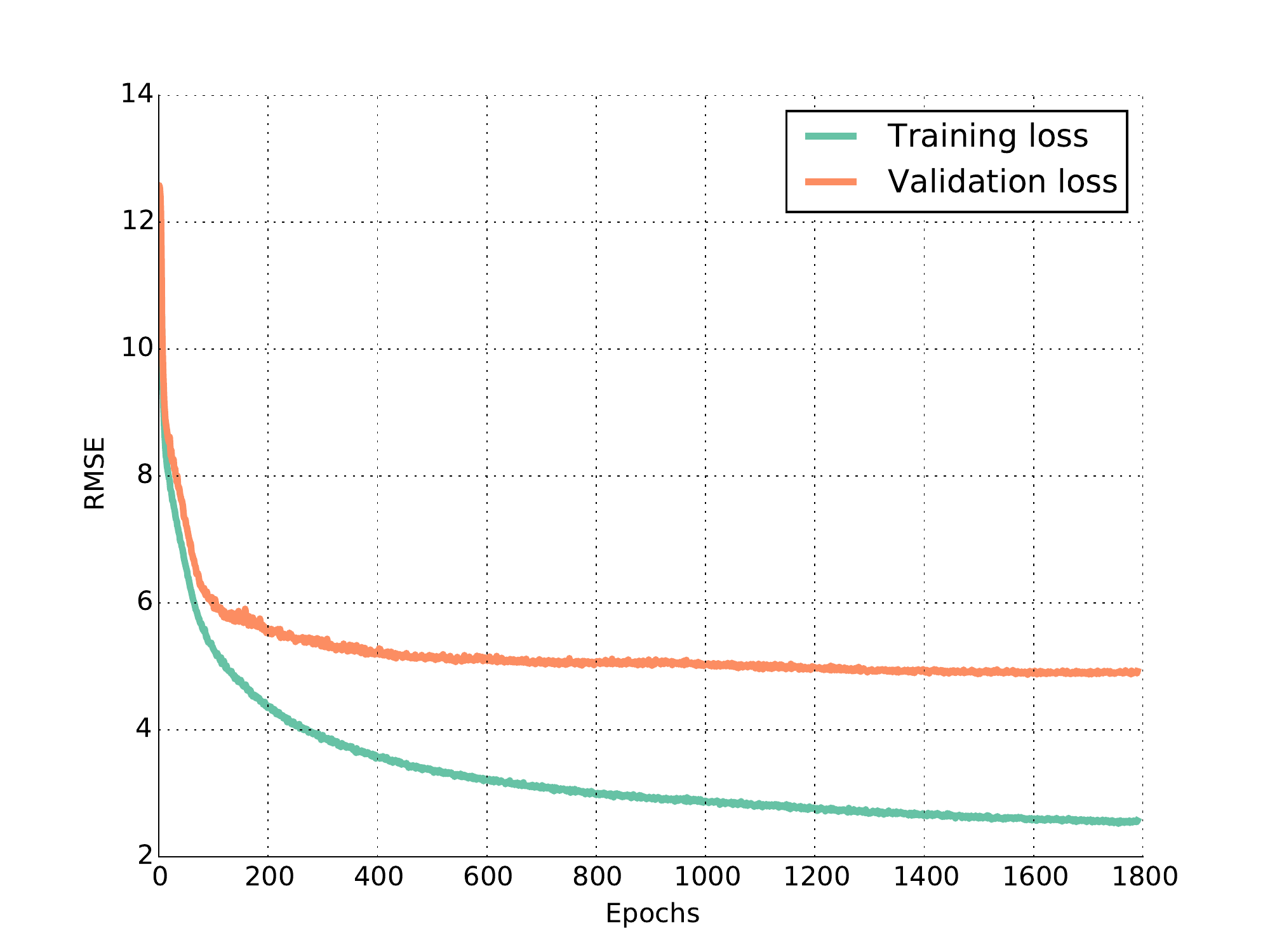}
\includegraphics[trim =0.0cm 0.0cm 0.0cm 0.0cm, clip = true, width=0.2\textwidth,height=0.25\textwidth]{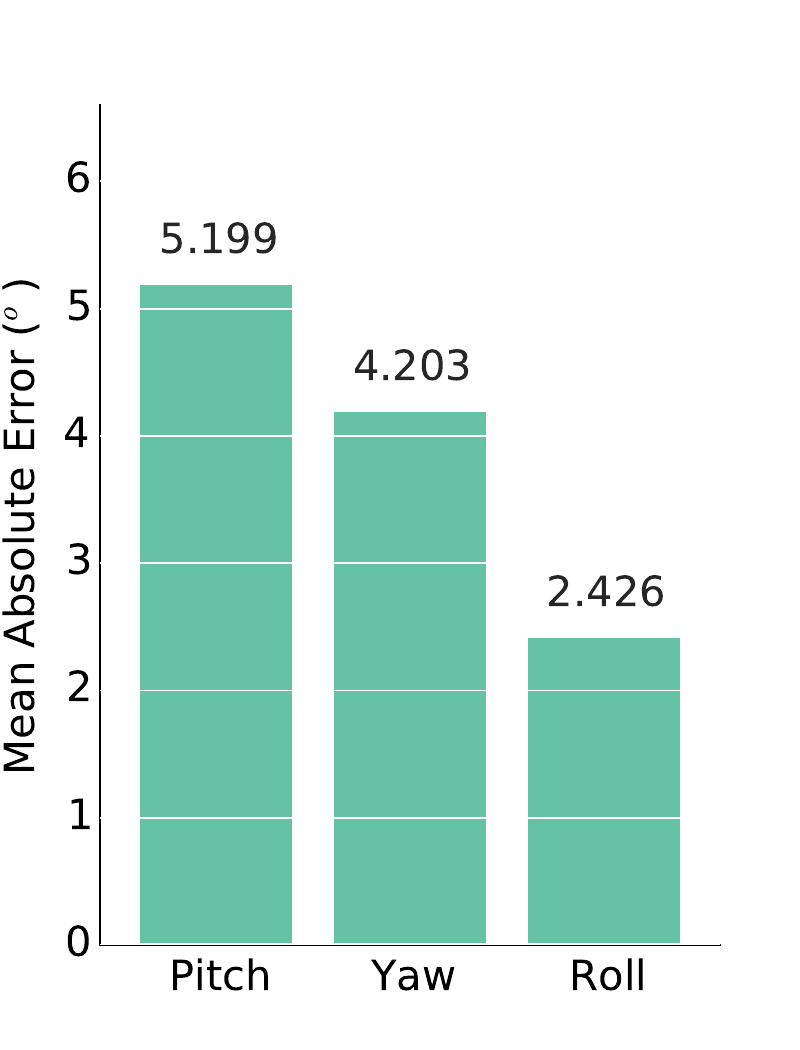}
\includegraphics[trim =0.0cm 0.0cm 0.0cm 0.0cm, clip = true, width=0.25\textwidth,height=0.25\textwidth]{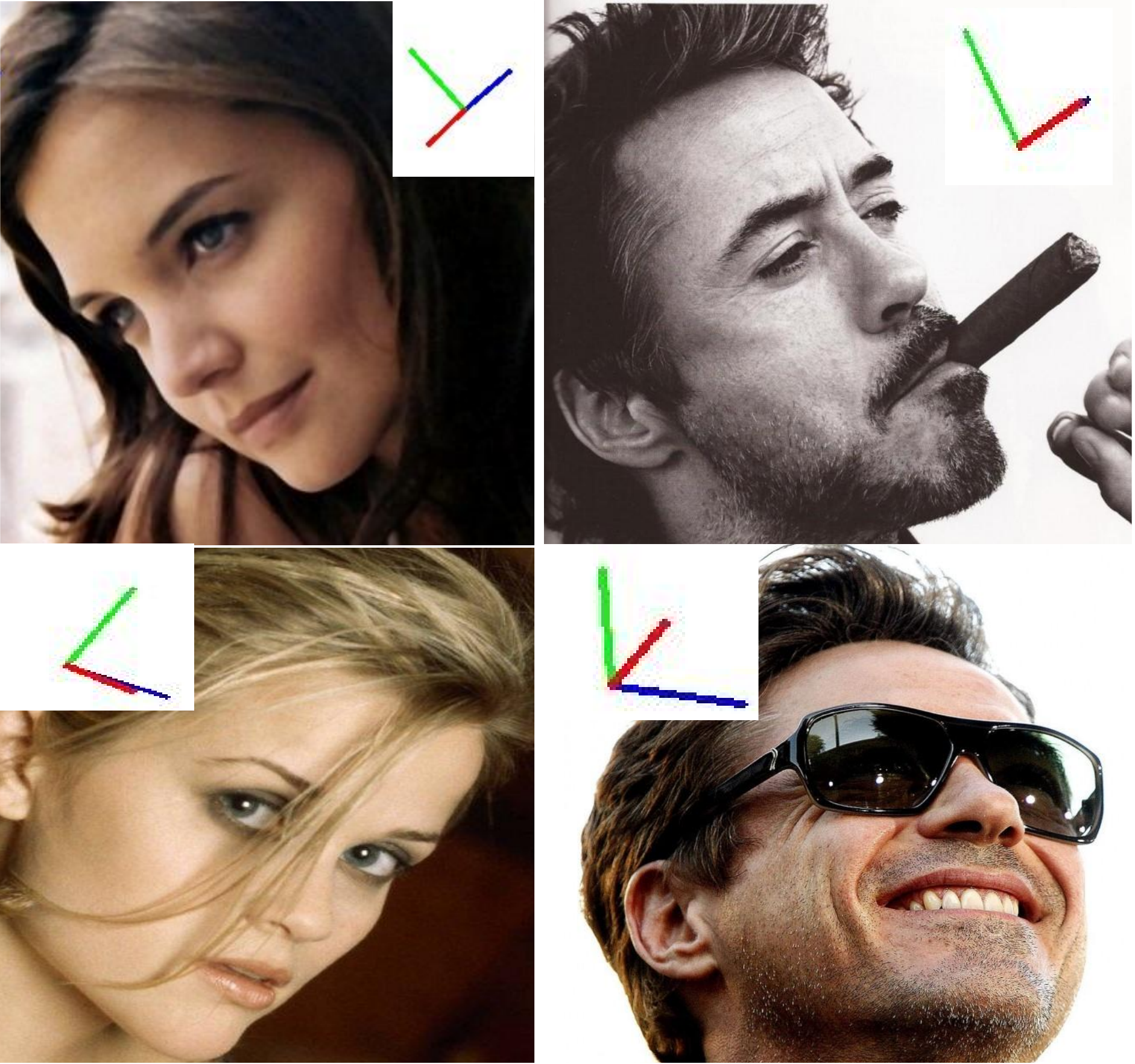}
\caption{Head pose estimation result. Left, learning curve of head pose network, with y axis the Root Mean Square Error (RMSE) and x axis the number of epochs; middle, absolute mean error on test set; right, example results of head pose estimation.}
\label{fig::headposeresult}
\end{figure*}
\subsection{Head  Pose Estimation}
We first evaluate the performance of head pose estimation. As we discussed before, it is very difficult to get the ground truth head pose for face images acquired in uncontrolled conditions. We calculate the pose based on the annotated facial landmark locations. We apply the trained deep ConvNet model on the test images of 300W and measure the performance.  The result is shown in Fig.~\ref{fig::headposeresult}. 
The absolute mean errors of the head pose pitch, yaw, roll angles are 5.1$^{\circ}$, 4.2$^{\circ}$ and 2.4$^{\circ}$, respectively. Some example results are shown on the right. Despite the work by Zhu \& Ramanan \cite{devacvpr2012face} is conceptually similar to our work in terms of simutaneuous head pose and facial landmarks estimation, we do not compare to it here because their work can only estimate very sparse head pose yaw angles (e.g. -15$^{\circ}$, 0$^{\circ}$ , 15 $^{\circ}$ ). 
\subsection{Face Alignment}
\begin{figure}
\includegraphics[trim =0.0cm 0.0cm .0cm 0.0cm, clip = true, width=0.48\textwidth,height=0.36\textwidth]{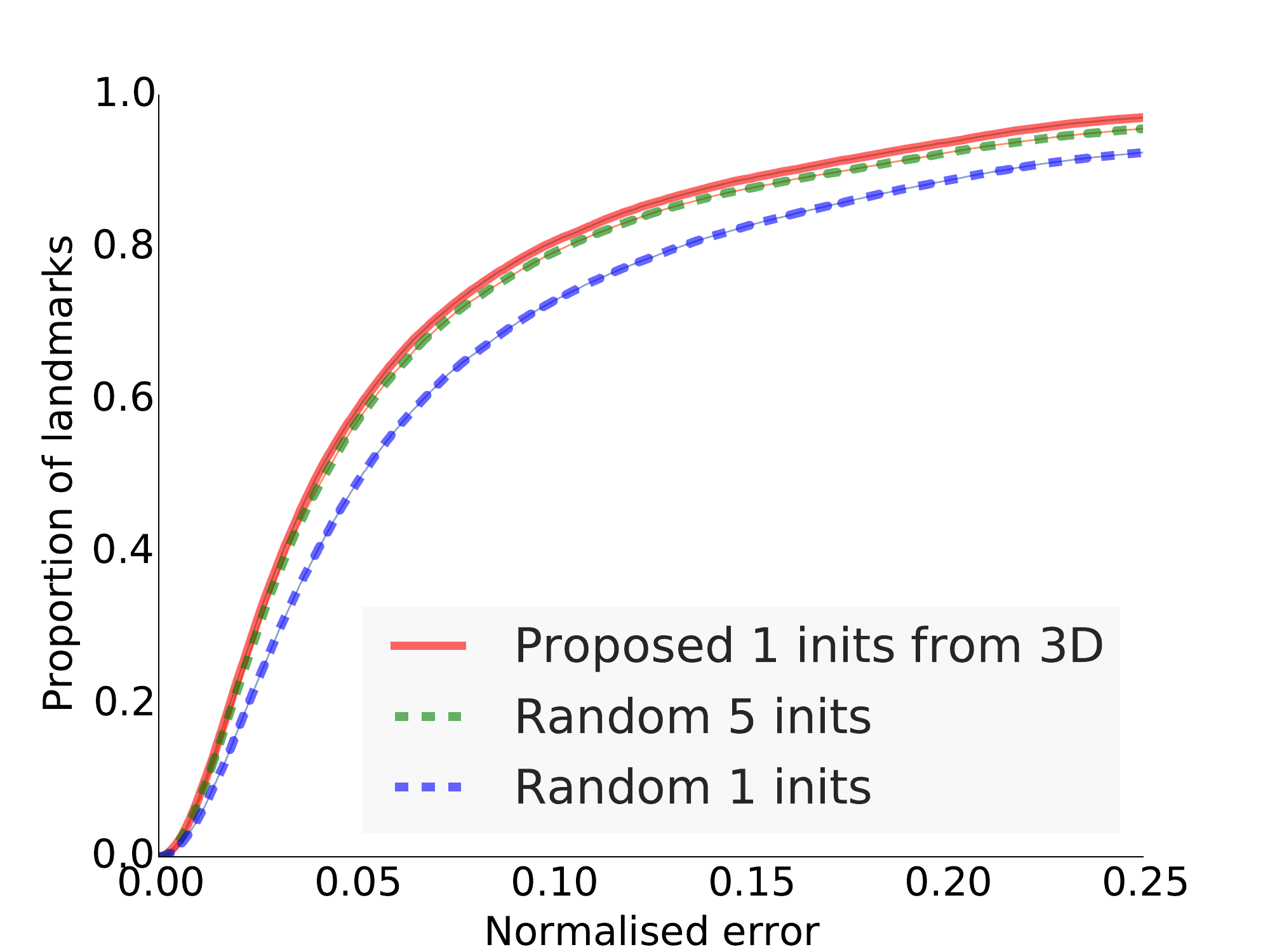}
\includegraphics[trim =0.0cm 0.0cm .0cm 0.0cm, clip = true, width=0.48\textwidth,height=0.36\textwidth]{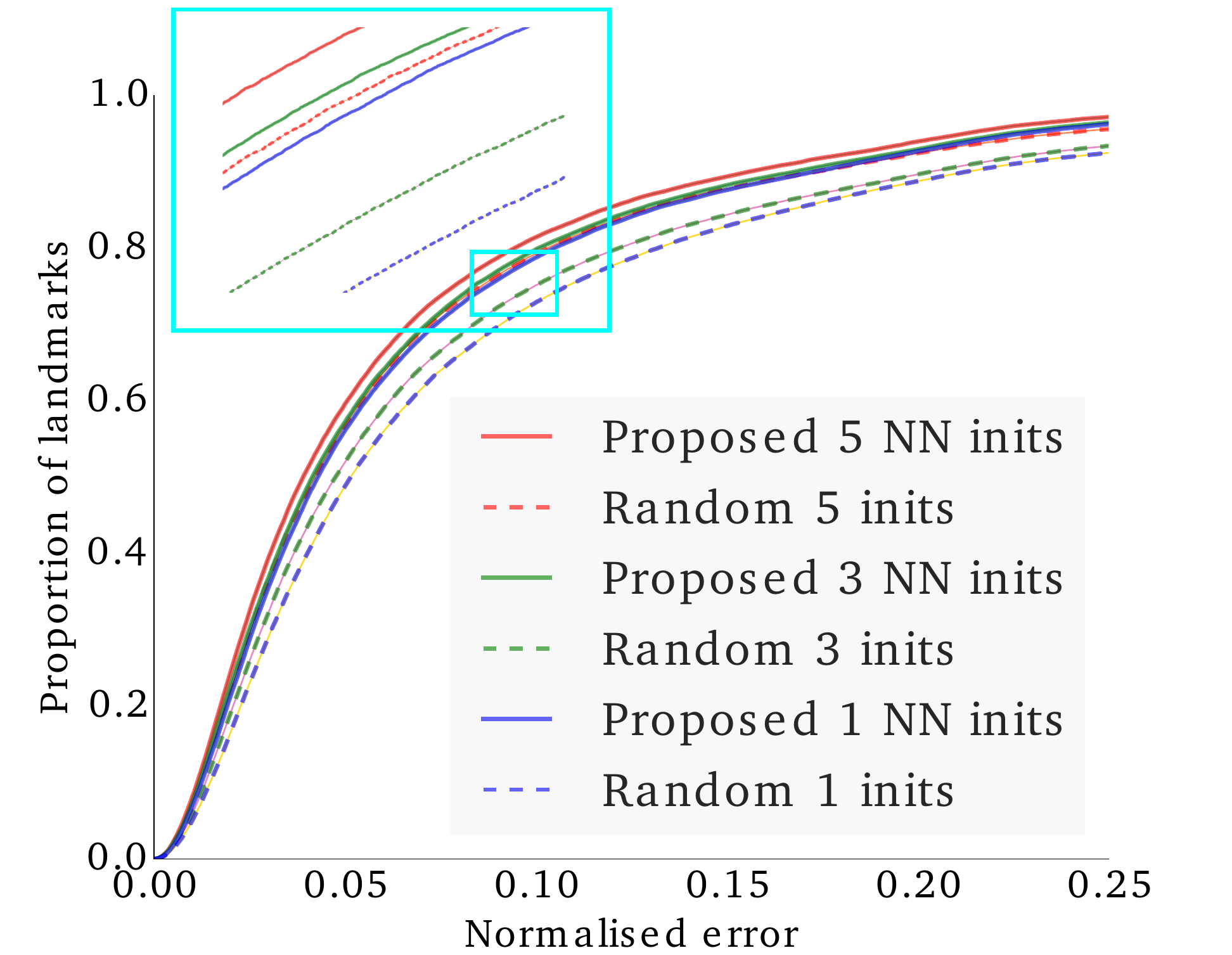}
\caption{Our proposed head pose based initialisation scheme vs. random initialisation scheme. Left, our 3D face shape based scheme; right, our Nearest Neighbour (NN) based scheme.}
\label{fig:comparewithbaseline}
\end{figure}

We first show the effectiveness of head pose based initialisation by comparing with the baseline strategy of the CPR framework \cite{suncvpr2012,burgos2013robust}, i.e., generating random initialisations from training samples. The comparison is shown in Fig.~\ref{fig:comparewithbaseline}. As can be seen on the left figure, by using one initialisation projected from 3D face shape, we obtain similar performance to the baseline approach with 5 initialisation shapes, and much better performance than that uses only one random initialisation shape. Similar superior performance is obtained by using nearest neighbour initialisation scheme, as shown on the right. By using more head pose based initialisations, we gain even better results, though the improvement is minor. It is worthy noting that by using our proposed initialisation scheme, we are able to decrease the number of failure cases (sample-wise average alignment error $>$ 0.1) from 130 to 69 (scheme 1) and \wenxuan{from 130} to 72 (scheme 2), nearly 50\%. Those samples are usually with large head pose variations and difficult for conventional face alignment methods. Moreover, by using one \wenxuan{set of} initialisation, the whole test procedure on one typical image takes 3.8 ms (0.3 ms for head pose estimation and 3.5 ms for cascaded face alignment). 

\begin{figure}
\includegraphics[trim =0.0cm 0.0cm .0cm 0.0cm, clip = true, width=0.48\textwidth,height=0.36\textwidth]{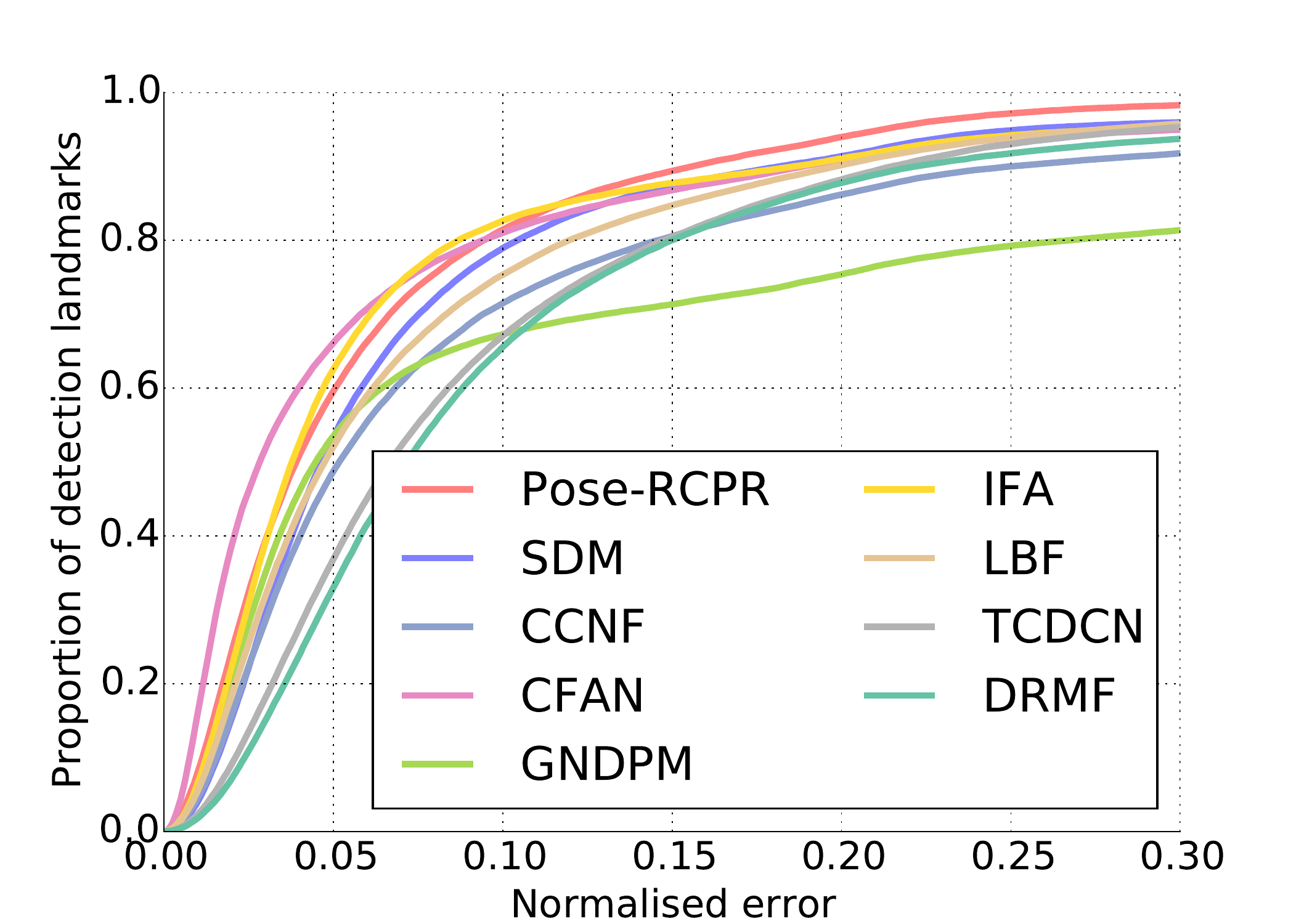}
\includegraphics[trim =0.0cm 0.0cm .0cm 0.0cm, clip = true, width=0.48\textwidth,height=0.36\textwidth]{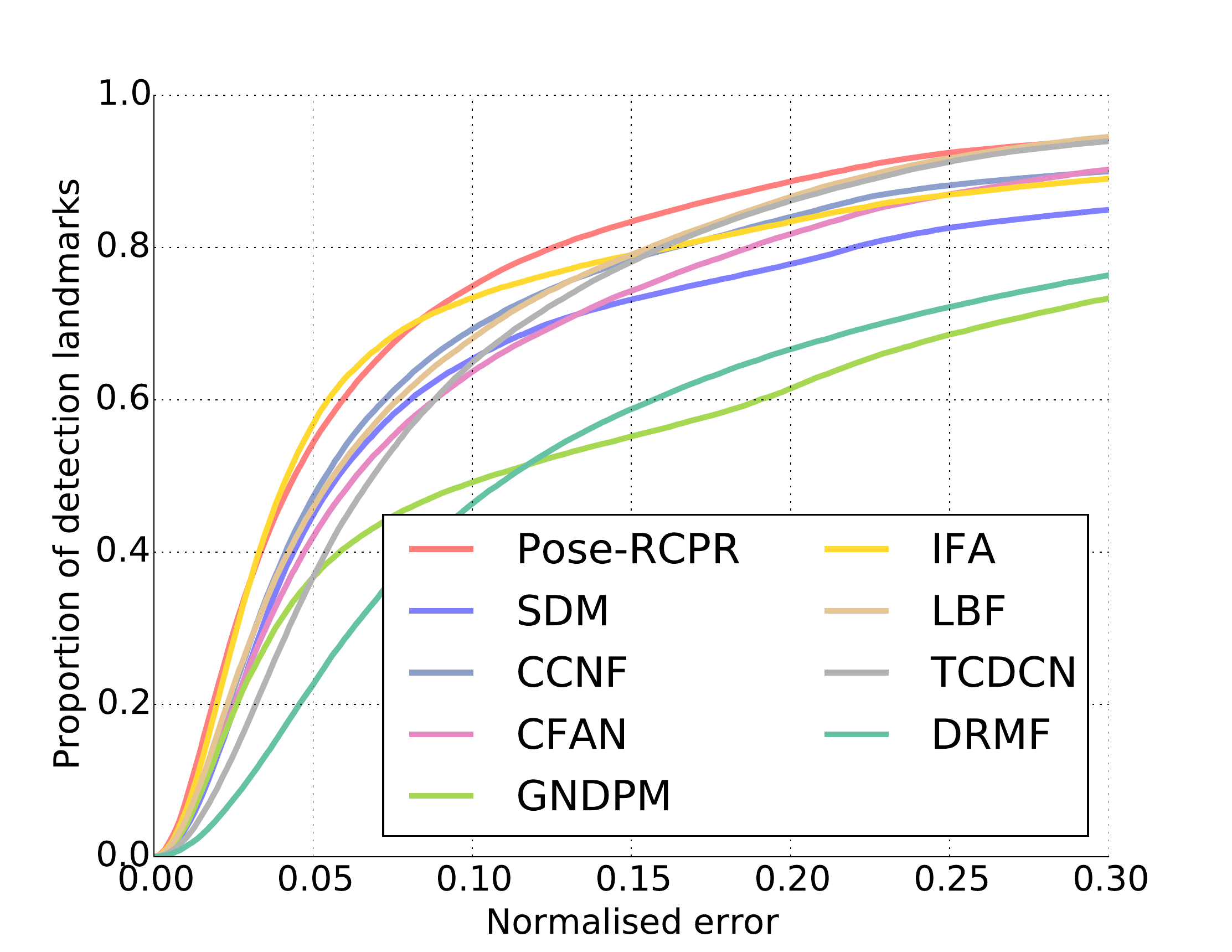}
\caption{Comparison with recent methods. Left, results from the \textit{best} face detection of each method; right, results from the common HeadHunter face detection.  Pose-RCPR is our proposed method using only 1 initialisation from 3D. }
\label{fig:comparewithsoa}
\end{figure}

We further compare the proposed method with recent state of the art methods including 5 holistic based methods (SDM \cite{xiong2013supervised}, IFA \cite{asthanaincremental}, LBF  \cite{renface}, CFAN \cite{cfaneccv2014}, TCDCN \cite{zhang2014facial}) and 3 local based methods (GNDPM \cite{tzimiropoulos2014gauss}, DRMF \cite{asthana2013robust}, CCNF \cite{baltruvsaitis2014continuous}). SDM and DRMF are trained using the Multi-PIE \cite{gross2010multi} dataset and detect 49 and 66 facial landmarks respectively. The rest of them are with models  trained on 300W datasets. When we run their model on the test images, we use the \textit{best} bounding boxes for a fair comparison. Best bounding box refers to Viola-Jones detection for SDM and RCPR and tight face detection provided by 300w dataset for the rest of them. The comparison is shown in Fig.~\ref{fig:comparewithsoa}. As can be seen, our proposed method shows competitive performance. We also compare the performance on another type of common face detection, HeadHunter, given its best performance in face detection. The result is shown on the right of Fig.~\ref{fig:comparewithsoa}. We observe that the performance of most methods deteriorate significantly when testing on HeadHunter face bounding boxes. Our method \wenxuan{provides most stable result}, despite the fact that the HeadHunter face bounding box is more overlapped with the face detection from 300W (both are tight boxes of facial landmarks) than with Viola-Jones face detection. We believe this robustness to face bounding box changes is partially due to our head pose based initialisation strategy. 

\section{Conclusion and Future Work}
In this paper we first \wenxuan{demonstrate} that most recent face alignment methods show failure cases when  large head pose variation is present. Based on the fact that cascaded face alignment is initialisation dependent, we proposed supervised initialisation schemes based on explicit head pose estimation. We use deep convolutional networks for head pose estimation and produce initialisation shape by either projecting a 3D face shape to the test image or searching nearest neighbour shapes from the training set. We demonstrated that using a more reliable initialisation is able to improve the face alignment performance with around 50\% failure decreasing. It also shows comparable or better performance when comparing to recent face alignment approaches. 

\wenxuan{Although we have managed to decrease} the failure cases to a certain degree, we have not fully solved this problem. There are several interesting directions for future research. First, using head pose based initialisation shapes in the training stage may further boost the performance. Second, we only test our method on RCPR, we believe the proposed scheme can be naturally applied to other  cascaded face alignment methods. It also raises several interesting questions. Do we need to make the cascaded learning model better for face alignment or to make the initialisation more reliable? Do we need more uniformly distributed data or a better model in order to make face alignment work better in wider range of head pose variations? We are going to investigate on these problem in our future research.  
\section*{Acknowledgement}
The work is sponsored by Cambridge VBRAD project from Jaguar-Land-Rover. We gratefully acknowledge NVIDIA for the donation of the Tesla GPU used for this research.
\bibliography{newbib}
\end{document}